\documentclass[letterpaper, 10 pt, conference]{ieeeconf}  % Comment this line out if you need a4paper
\pdfoutput=1
\IEEEoverridecommandlockouts                              % This command is only needed if
\usepackage{graphicx, subfigure, float, placeins}
\usepackage{amssymb, amsthm, amsfonts, amsmath, mathrsfs, textcomp, mathtools, gensymb, stmaryrd}
\usepackage{array, tabularx, booktabs, multirow}
\usepackage[font=small,labelfont=bf]{caption}
\usepackage{balance}

\usepackage{tikz, tikz-3dplot, pgfplots}
\usetikzlibrary{snakes, decorations, shapes, arrows, chains, positioning, shapes.symbols, patterns}
\pgfdeclarelayer{background}
\pgfdeclarelayer{foreground}
\pgfsetlayers{background,main,foreground}

\newcommand*\rot{\rotatebox{90}}
\newcolumntype{L}[1]{>{\raggedright\let\newline\\\arraybackslash\hspace{0pt}}m{#1}}
\newcolumntype{C}[1]{>{\centering\let\newline\\\arraybackslash\hspace{0pt}}m{#1}}
\newcolumntype{R}[1]{>{\raggedleft\let\newline\\\arraybackslash\hspace{0pt}}m{#1}}
\newcommand{\etal}{et~al.}

\usepackage{xcolor}
\definecolor{icra1}{HTML}{FDE728}   %(253 231  40)
\definecolor{icra2}{HTML}{C8E64C}   %(200 230  76)
\definecolor{icra3}{HTML}{8CD446}   %(140 212  70)
\definecolor{icra4}{HTML}{4DC742}   %( 77 199  66)
\definecolor{icra5}{HTML}{45D2B0}   %( 69 210 176)
\definecolor{icra6}{HTML}{438CCB}   %( 67 140 203)
\definecolor{icra7}{HTML}{4262FE}   %( 66  98 254)
\definecolor{icra8}{HTML}{5240C3}   %( 82  64 195)
\definecolor{icra9}{HTML}{7C3FC0}   %(124  63 192)
\definecolor{icra10}{HTML}{D145C1}  %(209  69 193)
\definecolor{icra11}{HTML}{FF5454}  %(255  84  84)
\definecolor{icra12}{HTML}{FF8000}  %(255 128   0)
\definecolor{icra13}{HTML}{FFA054}  %(255 160  84)

\definecolor{cnn0}{HTML}{440154}%{404788}
\definecolor{cnn1}{HTML}{404788}%{404788}
\definecolor{cnn2}{HTML}{20A387}
\definecolor{cnn3}{HTML}{287D8E}

\definecolor{colorcnn}{HTML}{440154}%{404788}
\definecolor{colorwarp}{HTML}{404788}%{404788}

\newcommand{\block}[2]{\rot{\shortstack[l]{\tikz \draw[color=black, fill=#1] (0,0) rectangle ++(0.25,0.25) {}; \\ {#2}}}}

\makeatletter
\let\NAT@parse\undefined
\makeatother

\usepackage{footnote}

\usepackage[bookmarks=false, pdfauthor={summer.icequeen}, colorlinks=true, linkcolor=blue, citecolor=red, filecolor=cyan, urlcolor=blue]{hyperref} % should go last
\graphicspath{{./figures/}}

\title{\Large \bf Multi-View Deep Learning for Consistent Semantic Mapping with RGB-D Cameras}
\author{Lingni Ma, J{\"o}rg St\"uckler, Christian Kerl and Daniel Cremers%
\thanks{Authors are with Computer Vision and Artificial Intelligence Group, Department of Computer Science, Technical University of Munich, ({\tt\small \{lingni,stueckle,kerl,cremers\}@in.tum.de}). } %
\thanks{This work is accepted by International Conference on Intelligent Robots and Systems, 2017. It is funded by ERC Consolidator Grant 3D Reloaded (649323).}%
}

\begin{document}
\maketitle

\thispagestyle{empty}
\pagestyle{empty}

%%%%%%%%%%%%%%%%%%%%%%%%%%%%%%%%%%%%%%%%%%%%%%%%%%%%%%%%%%%%%%%%%%%%%%%%%%%%%%%%
\begin{abstract}
Visual scene understanding is an important capability that enables robots to purposefully act in their environment. In this paper, we propose a novel deep neural network approach to predict semantic segmentation from RGB-D sequences. The key innovation is to train our network to predict multi-view consistent semantics in a self-supervised way. At test time, its semantics predictions can be fused more consistently in semantic keyframe maps than predictions of a network trained on individual views. We base our network architecture on a recent single-view deep learning approach to RGB and depth fusion for semantic object-class segmentation and enhance it with multi-scale loss minimization. We obtain the camera trajectory using RGB-D SLAM and warp the predictions of RGB-D images into ground-truth annotated frames in order to enforce multi-view consistency during training. At test time, predictions from multiple views are fused into keyframes. We propose and analyze several methods for enforcing multi-view consistency during training and testing. We evaluate the benefit of multi-view consistency training and demonstrate that pooling of deep features and fusion over multiple views outperforms single-view baselines on the NYUDv2 benchmark for semantic segmentation. Our end-to-end trained network achieves state-of-the-art performance on the NYUDv2 dataset in single-view segmentation as well as multi-view semantic fusion.
%For example, for 13-class segmentation we obtain 79.13\% global pixelwise accuracy, 70.59\% average classwise accuracy and 59.07\% IoU scores.
\end{abstract}

%%%%%%%%%%%%%%%%%%%%%%%%%%%%%%%%%%%%%%%%%%%%%%%%%%%%%%%%%%%%%%%%%%%%%%%%%%%%%%%% introduction
\section{Introduction}
Intelligent robots require the ability to understand their environment through parsing and segmenting the 3D scene into meaningful objects. The rich appearance-based information contained in images renders vision a primary sensory modality for this task.

\begin{figure}[t!]

    \centering
    \begin{tikzpicture}[inner sep=1pt]

    \tikzstyle{cmt}=[font=\footnotesize,text=black]
    \tikzstyle{arr}=[color=black, <-,>=stealth', line width=0.6pt, line cap=rounded] %, rounded corners
    \tikzstyle{lsty}=[color=black, line width=1pt, line cap=rounded]
    \def\hy{18mm}
    \def\hx{29mm}
    % \defcolorcnn{cnn0}
    % \defcolorwarp{cnn1}
    \def\cnn#1#2
    {
    \begin{scope}[xshift=#1, yshift=#2]
        \draw[lsty,color=colorcnn, fill=white] (0,0) --++(0,\hy)--++(\hx, -\hx*0.1)--++(\hx, \hx*0.1) --++(0,-\hy) --++(-\hx, \hx*0.1) --++(-\hx, -\hx*0.1) --cycle;
    \end{scope}
    }

    \def\mvblock#1#2#3
    {
        \def\xx{0.9/#3}
        \draw[lsty, color=colorwarp, fill=none](#1,#2)--++(0,-\xx/2)--++(\xx,0)--++(0,\xx)--++(-\xx,0)--++(0,-\xx/2)--cycle;
    }

    \tikzstyle{tblock} = [draw, rectangle, rounded corners, thick, text centered, font=\footnotesize, minimum height=4mm, color=colorwarp,]

    \begin{scope}%[yshift=1.5cm]
        \cnn{0}{0mm}
        \node(p1)at(-0.1,\hy/2)[anchor=south east]{\includegraphics[width=1.2cm]{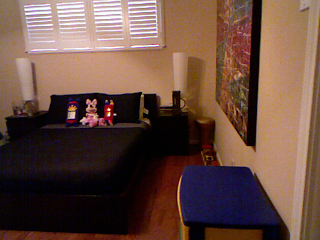}};
        \node(p2)at(p1.south)[anchor=north]{\includegraphics[width=1.2cm]{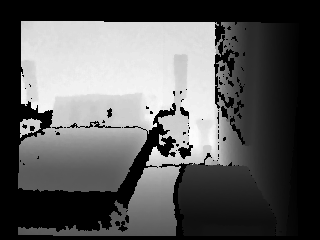}};
        \node at(\hx, \hy*0.5)[yshift=0mm] {CNN $f(\mathcal{I}_i, \mathcal{W})$};
        \node(t11) at (p2.south) [anchor=north, font=\footnotesize, text centered, yshift=-4pt]
        {view $\mathcal{I}_i$};
        % \node(t12) at(t11.north east) [anchor=north west, cmt, text=colorcnn, text width=18mm, text centered, xshift=60mm,yshift=1mm]
        % {warped outputs};
        %  $\mathcal{F}_i\big(\omega(\mathbf{x}, \boldsymbol{\xi}_i)\big)$};

        \node(i11) at(p2.south east) [anchor=north west,shift={(1.8, -0.1)}] {\includegraphics[width=0.3cm]{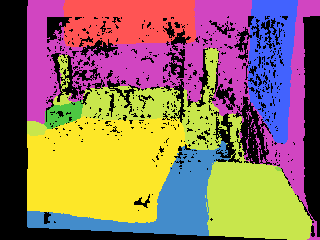}};
        \node(i12) at(i11.north east) [anchor=north west, xshift=4pt]{\includegraphics[width=0.4cm]{40_warped.png}};
        \node(i13) at(i12.north east) [anchor=north west, xshift=4pt]{\includegraphics[width=0.6cm]{40_warped.png}};
        \node(i14) at(i13.north east) [anchor=north west, xshift=5pt]{\includegraphics[width=0.8cm]{40_warped.png}};
        \node(i15) at(i14.north east) [anchor=north west, xshift=5pt]{\includegraphics[width=1.0cm]{40_warped.png}};
        \node(i16) at(p1.south east) [anchor= west, xshift=60mm, yshift=0mm] {\includegraphics[width=1.2cm]{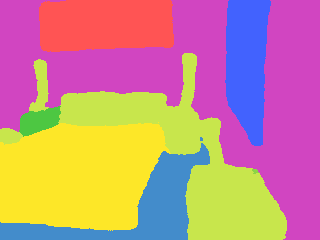}};
        \node(t21) at(i15.east) [anchor=west, cmt, text=colorcnn, text width=10mm, text centered, xshift=1mm, yshift=-0mm]
        {warped outputs};

        \draw[color=colorcnn, -triangle 60,line width=1.5pt, postaction={draw, line width=3.5pt, shorten >=1mm, -}] (3.6,0.2)--++(0,-0.4);
        \draw[color=colorwarp, -triangle 60,line width=1.5pt, postaction={draw, line width=3.5pt, shorten >=1mm, -}] (3.6,-1.0)--++(0,-0.4);

        % \node at(i11.west)[anchor=west, tblock, yshift=-10.5mm, text width=41mm,inner sep=1pt] {multi-view consistency supervision};

        \node(i21) at(i11.south)[anchor=north, yshift=-27mm]{\includegraphics[width=0.3cm]{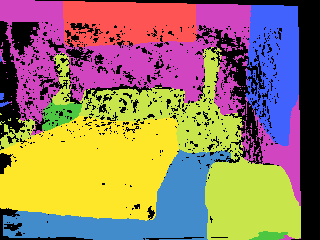}};
        \node(i22) at(i21.south east) [anchor=south west, xshift=4pt]{\includegraphics[width=0.4cm]{66_warped.png}};
        \node(i23) at(i22.south east) [anchor=south west, xshift=4pt]{\includegraphics[width=0.6cm]{66_warped.png}};
        \node(i24) at(i23.south east) [anchor=south west, xshift=5pt]{\includegraphics[width=0.8cm]{66_warped.png}};
        \node(i25) at(i24.south east) [anchor=south west, xshift=5pt]{\includegraphics[width=1.0cm]{66_warped.png}};
        \node(i26) at(i16.south) [anchor=north, yshift=-45mm] {\includegraphics[width=1.2cm]{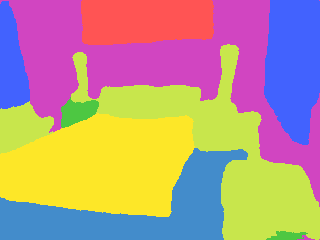}};

        \node(t22) at(i25.east) [anchor=west, cmt, text=colorcnn, text width=10mm, text centered, xshift=1mm, yshift=-0mm]
        {warped outputs};

        \draw[color=colorcnn, -triangle 60,line width=1.5pt, postaction={draw, line width=3.5pt, shorten >=1mm, -}]
        (3.6, -3.9)--++(0,0.4);
        \draw[color=colorwarp, -triangle 60,line width=1.5pt, postaction={draw, line width=3.5pt, shorten >=1mm, -}]
        (3.6, -2.7)--++(0,0.4);

        \node(i31) at(i11.south)[anchor=north, yshift=-12mm]{\includegraphics[width=0.3cm]{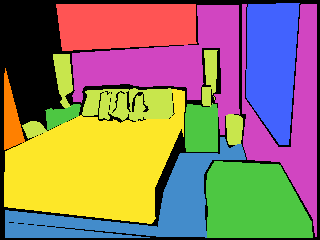}};
        \node(i32) at(i31.east) [anchor=west, xshift=4pt]{\includegraphics[width=0.4cm]{280gt.png}};
        \node(i33) at(i32.east) [anchor=west, xshift=4pt]{\includegraphics[width=0.6cm]{280gt.png}};
        \node(i34) at(i33.east) [anchor=west, xshift=5pt]{\includegraphics[width=0.8cm]{280gt.png}};
        \node(i35) at(i34.east) [anchor=west, xshift=5pt]{\includegraphics[width=1.0cm]{280gt.png}};
        % \node(i36) at(i16.south) [anchor=north, yshift=-36mm] {\includegraphics[width=1.2cm]{66_current.png}};
        \node(t32) at(i31.west) [anchor=east, cmt, text=black, text width=18mm, text centered, xshift=-1mm,yshift=0mm]
        {reference-view groundtruth};

        \node(cc1) at(t32.north) [anchor=south, cmt, text=colorwarp, text width=18mm, text centered, xshift=20mm,yshift=0mm]
        {consistency supervision};

        \node(cc2) at(t32.south) [anchor=north, cmt, text=colorwarp, text width=18mm, text centered, xshift=20mm,yshift=0mm]
        {consistency supervision};

    \end{scope}

    \begin{scope}[yshift=-5.5cm]
        \cnn{0}{0mm}
        \node(p3)at(-0.1,\hy/2)[anchor=south east]{\includegraphics[width=1.2cm]{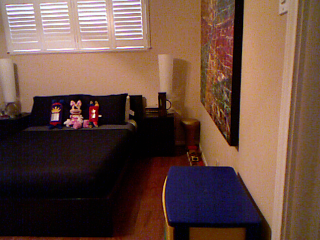}};
        \node(p4)at(p3.south)[anchor=north]{\includegraphics[width=1.2cm]{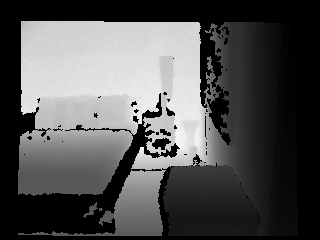}};
        \node at(\hx, \hy*0.5)[yshift=0mm] {CNN $f(\mathcal{I}_j, \mathcal{W})$};
        \node(t21) at (p3.north) [anchor=south, font=\footnotesize, text centered, yshift=4pt]
        {view $\mathcal{I}_j$};
        % \node(t22) at(t21.south east) [anchor=south west, cmt, text=colorcnn, text width=12mm, text centered, xshift=60mm, yshift=-1mm]
        % {warped outputs};
        %  $\mathcal{F}_j\big(\omega(\mathbf{x}, \boldsymbol{\xi}_j)\big)$};
    \end{scope}

\end{tikzpicture}
\caption{We train our CNN to predict multi-view consistent semantic segmentations for RGB-D images. The key innovation is to enforce consistency by warping CNN feature maps from multiple views into a common reference view using the SLAM trajectory and to supervise training at multiple scales. Our approach improves performance for single-view segmentation and is specifically beneficial for multi-view fused segmentation.}
\label{fig:teaser}
\end{figure}

In recent years, large progress has been achieved in semantic segmentation of images. Most current state-of-the-art approaches apply deep learning for this task. With RGB-D cameras, appearance as well as shape modalities can be combined to improve the semantic segmentation performance. Less explored, however, is the usage and fusion of multiple views onto the same scene which appears naturally in the domains of 3D reconstruction and robotics. Here, the camera is moving through the environment and captures the scene from multiple view points. Semantic SLAM aims at aggregating several views in a consistent 3D geometric and semantic reconstruction of the environment.

In this paper, we propose a novel deep learning approach for semantic segmentation of RGB-D images with multi-view context. We base our network on a recently proposed deep convolutional neural network (CNN) for RGB and depth fusion~\cite{lingni16accv} and enhance the approach with multi-scale deep supervision. Based on the trajectory obtained through RGB-D simultaneous localization and mapping (SLAM), we further regularize the CNN training with multi-view consistency constraints as shown in Fig.~\ref{fig:teaser}. We propose and evaluate several variants to enforce multi-view consistency during training. A shared principle is using the SLAM trajectory estimate to warp network outputs of multiple frames into the reference view with ground-truth annotation. By this, the network not only learns features that are invariant under view-point change. Our semi-supervised training approach also makes better use of the annotated ground-truth data than single-view learning. This alleviates the need for large amounts of annotated training data which is expensive to obtain. Complementary to our training approach, we aggregate the predictions of our trained network in keyframes to increase segmentation accuracy at testing. The predictions of neighboring images are fused into the keyframe based on the SLAM estimate in a probabilistic way.

In experiments, we evaluate the performance gain achieved through multi-view training and fusion at testing over single-view approaches. Our results demonstrate that multi-view max-pooling of feature maps during training best supports multi-view fusion at testing. Overall we find that enforcing multi-view consistency during training significantly improves fusion at test time versus fusing predictions from networks trained on single views. Our end-to-end training achieves state-of-the-art performance on the NYUDv2 dataset in single-view segmentation as well as multi-view semantic fusion. While the fused keyframe segmentation can be directly used in robotic perception, our approach can also be useful as a building block for semantic SLAM using RGB-D cameras.

\begin{figure*}
    \centering
    \begin{tikzpicture}

    \tikzstyle{cmt}=[font=\footnotesize,text=black, text centered]
    \tikzstyle{arr}=[color=black, ->,>=stealth', line width=0.8pt, line cap=rounded, rounded corners]

    \def\dec{cnn0}
    \def\decwarp{cnn1}
    \def\colorwarp{cnn1}
    \def\encrgb{cnn2}
    \def\encdep{cnn3}

    \def\warpLayer#1#2#3#4
    {
    \begin{scope}[xshift=#3,yshift=#4]
    \draw[draw, line width=0.8pt, line cap=round, fill=none] (0,0, 0) -- ++(#1,0,0) -- ++(0,#2,0) -- ++(-#1,0,0) -- cycle;
    \draw[draw, line width=0.8pt, line cap=round, fill=none] (#1,0, 0) -- ++(0,0,-#2) -- ++(0,#2,0) -- ++(0,0,#2) -- cycle;
    \draw[draw, line width=0.8pt, line cap=round, fill=none] (0,#2, 0) -- ++(0,0,-#2) -- ++(#1,0,0) -- ++(0,0,#2) -- ++(-#1,0,0) -- cycle;
    \end{scope}
    }

    \def\oneLayer#1#2#3#4
    {
    \begin{scope}[xshift=#3,yshift=#4]
    \def\xx{#1}
    \def\yz{#2/4}
    \draw[draw, dashed, line width=0.3pt, line cap=round, fill=none] (0,\yz*1.5, -\yz*2) -- ++(\xx,0,0) -- ++(0,\yz,0) -- ++(-\xx,0,0) -- cycle;
    \draw[draw, dashed, line width=0.3pt, line cap=round, fill=none] (\xx,\yz*1.5, -\yz*2) -- ++(0,0,-\yz) -- ++(0,\yz,0) -- ++(0,0,\yz) -- cycle;
    \draw[draw, dashed, line width=0.3pt, line cap=round, fill=none] (0,\yz*2.5, -\yz*2) -- ++(0,0,-\yz) -- ++(\xx,0,0) -- ++(0,0,\yz) -- ++(-\xx,0,0) -- cycle;

    \draw[draw, line width=0.8pt, line cap=round, fill=none] (0,0, 0) -- ++(#1,0,0) -- ++(0,#2,0) -- ++(-#1,0,0) -- cycle;
    \draw[draw, line width=0.8pt, line cap=round, fill=none] (#1,0, 0) -- ++(0,0,-#2) -- ++(0,#2,0) -- ++(0,0,#2) -- cycle;
    \draw[draw, line width=0.8pt, line cap=round, fill=none] (0,#2, 0) -- ++(0,0,-#2) -- ++(#1,0,0) -- ++(0,0,#2) -- ++(-#1,0,0) -- cycle;
    \end{scope}
    }

    % rgb encoder
    \begin{scope}[yshift=26mm, color=\encrgb]
        \oneLayer{0.1}{0.70mm}{0}{0}
        \oneLayer{0.4}{0.50mm}{10mm}{4mm}
        \oneLayer{0.8}{0.40mm}{21mm}{6mm}
        \oneLayer{1.1}{0.20mm}{35mm}{10mm}
        \oneLayer{1.2}{0.12mm}{50mm}{11mm}

        % \node at (-1.0, 2.8)[cmt] {RGB views};
        \node at (0.5, 3.0) [cmt] {conv1};
        \node at (1.6, 2.7) [cmt] {conv2};
        \node at (2.8, 2.4) [cmt] {conv3};
        \node at (4.2, 2.0) [cmt] {conv4};
        \node at (5.7, 1.8) [cmt] {conv5};
        \node at (4.0, 0.1) [cmt,color=black, text width=18mm, color=\encdep] {RGB-D fusion $\mathcal{F}_{rgb} \oplus \mathcal{F}_{d}$};

    \end{scope}
    % depth
    \begin{scope}[color=\encdep]
        \oneLayer{0.1}{0.70mm}{0}{0}
        \oneLayer{0.4}{0.50mm}{10mm}{4mm}
        \oneLayer{0.8}{0.40mm}{21mm}{6mm}

        % \node at (-1.2,-0.) [cmt] {depth views};
        \node at (0.3,-0.3) [cmt] {conv1-d};
        \node at (1.4,-0.0) [cmt] {conv2-d};
        \node at (2.7, 0.2) [cmt] {conv3-d};

        \draw[arr,color=\encdep] (0.5, 2.4) [out=60,in=300] to (0.5, 3.0);
        \draw[arr,color=\encdep] (1.5, 2.0) [out=60,in=300] to (1.5, 3.1);
        \draw[arr,color=\encdep] (2.7, 1.9) [out=60,in=300] to (2.7, 3.2);
    \end{scope}

    % warping and consistency supervision
    \begin{scope}[color=\decwarp, xshift=67mm]
      \oneLayer{1.2}{0.12mm}{0mm}{11mm}
      \oneLayer{1.1}{0.20mm}{15mm}{10mm}
      \oneLayer{0.8}{0.40mm}{30mm}{6mm}
      \oneLayer{0.4}{0.50mm}{45mm}{4mm}
      \oneLayer{0.1}{0.70mm}{58mm}{0mm}

      % \node at (0.7, 0.8) [cmt] {MVCS5};
      % \node at (2.1, 0.6) [cmt] {MVCS4};
      % \node at (3.6, 0.2) [cmt] {MVCS3};
      % \node at (5.0, 0.0) [cmt] {MVCS2};
      % \node at (6.2,-0.3) [cmt] {MVCS1};

      \tikzstyle{tblock} = [draw, rectangle, rounded corners, thick, text centered, font=\footnotesize, minimum height=4mm, color=\colorwarp,]
      \node at(3.3,0.8)[anchor=north, tblock, yshift=-10.5mm, text width=66mm,inner sep=1pt] {multi-view consistency supervision};

      \draw[arr, <-, color=\decwarp] (0.7, -0.3) --++(0,1.4);
      \draw[arr, <-, color=\decwarp] (2.1, -0.3) --++(0,1.3);
      \draw[arr, <-, color=\decwarp] (3.6, -0.3) --++(0,0.9);
      \draw[arr, <-, color=\decwarp] (5.0, -0.3) --++(0,0.8);
      \draw[arr, <-, color=\decwarp] (6.2, -0.3) --++(0,0.6);

    \end{scope}
    % # decoder
    \begin{scope}[xshift=67mm, yshift=26mm, color=\dec]
        \oneLayer{1.2}{0.12mm}{0mm}{11mm}
        \oneLayer{1.1}{0.20mm}{15mm}{10mm}
        \oneLayer{0.8}{0.40mm}{30mm}{6mm}
        \oneLayer{0.4}{0.50mm}{45mm}{4mm}
        \oneLayer{0.1}{0.70mm}{58mm}{0mm}

        \node at (0.7, 1.8) [cmt] {deconv5};
        \node at (2.2, 2.0) [cmt] {deconv4};
        \node at (3.7, 2.4) [cmt] {deconv3};
        \node at (5.0, 2.7) [cmt] {deconv2};
        \node at (6.2, 3.0) [cmt] {deconv1};
        % \node at (7.6, 2.8) [cmt, text width=1.5cm, text centered] {keyframe segmentation};

        \draw[arr,color=\dec] (0.6, 1.1)[out=300,in=60] to (0.6, -1.0)
        node at (0.0,0.1)[cmt, text width=18mm, color=\dec]{warping into referene view $\mathcal{F}\big(\omega(\mathbf{x}, \boldsymbol{\xi})\big)$};
        \draw[arr,color=\dec] (2.2, 1.0)[out=300,in=60] to (2.2, -0.8);
        \draw[arr,color=\dec] (3.6, 0.6)[out=300,in=60] to (3.6, -0.4);
        \draw[arr,color=\dec] (5.1, 0.5)[out=300,in=60] to (5.1, -0.2);
        \draw[arr,color=\dec] (6.3, 0.4)[out=300,in=60] to (6.3, -0.1);
    \end{scope}

    \node(rgb1)  at (-1.3, 4.5) {\includegraphics[width=1.2cm]{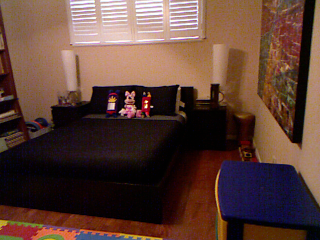}};%{219rgb.png}};
    \node(rgb2)  at (-1.1, 4.0) {\includegraphics[width=1.2cm]{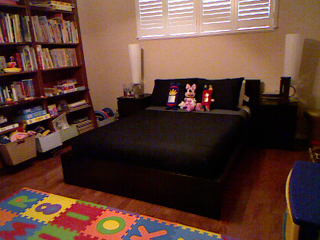}};%{35rgb.png}};
    \node(rgb3)  at (-0.9, 3.5) {\includegraphics[width=1.2cm]{rgb_frame_40.png}};%{413_rgb.png}};
    \node at (rgb2.north)[cmt, anchor=south, yshift=5mm, text width=16mm] {RGB views};

    \node(dimg1) at (-1.3, 2.1) {\includegraphics[width=1.2cm]{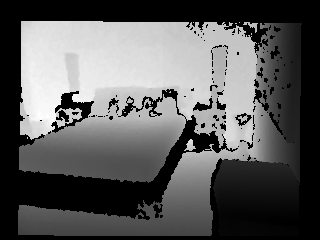}}; %{219d.png}};
    \node(dimg2) at (-1.1, 1.6) {\includegraphics[width=1.2cm]{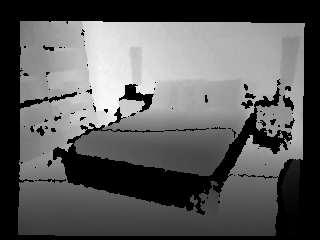}}; %{35d.png}};
    \node(dimg3) at (-0.9, 1.1) {\includegraphics[width=1.2cm]{d_frame_40.png}}; %{413_d.png}};
    \node at (dimg2.south)[cmt, anchor=north, yshift=-5mm, text width=16mm] {depth views};

    \node(pred1) at(14.2, 4.5) {\includegraphics[width=1.2cm]{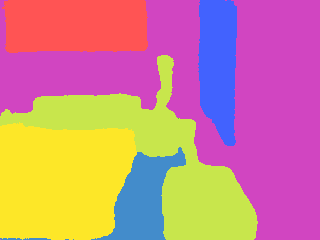}};
    \node(pred2) at(14.4, 4.0) {\includegraphics[width=1.2cm]{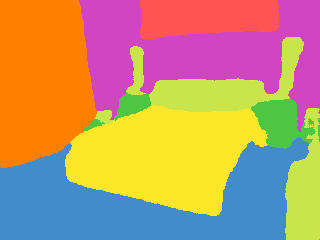}};
    \node(pred3) at(14.6, 3.5) {\includegraphics[width=1.2cm]{40_current.png}};
    \node at (pred2.north)[cmt, anchor=south, yshift=5mm, text width=16mm] {semantic segmentation};

    \node(warp1) at(14.2, 2.1) {\includegraphics[width=1.2cm]{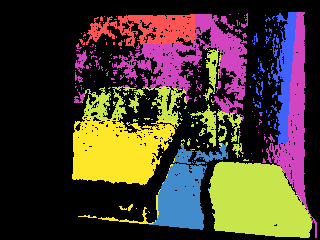}};
    \node(warp2) at(14.4, 1.6) {\includegraphics[width=1.2cm]{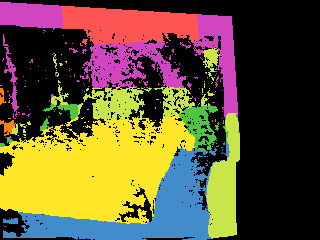}};
    \node(warp3) at(14.6, 1.1) {\includegraphics[width=1.2cm]{40_warped.png}};
    \node at (warp2.south)[cmt, anchor=north, yshift=-5mm, text width=16mm] {warped outputs};

    \end{tikzpicture}

    \caption{The CNN encoder-decoder architecture used in our approach. Input to the network are RGB-D sequences with corresponding poses from SLAM trajectory. The encoder contains two branches to learn features from RGB-D data as inspired by FuseNet~\cite{lingni16accv}. The obtained low-resolution high-dimension feature maps are successively refined through deconvolutions in the decoder. We warp feature maps into a common reference view and enforce multi-view consistency with various constraints. The network is trained in a deeply-supervised manner where loss is computed at all scales of the decoder.}
    \label{fig:network}

\end{figure*}

\section{Related Work}
% \subsection{CNN related semantic segmentation}

Recently, remarkable progress has been achieved in semantic image segmentation using deep neural networks and, in particular, CNNs. On many benchmarks, these approaches excell previous techniques by a great margin.

{\bf Image-based Semantic Segmentation.} As one early attempt, Couprie~\etal~\cite{couprie13iclr} propose a multiscale CNN architecture to combine information at different receptive field resolutions and achieved reasonable segmentation results. Gupta~\etal~\cite{gupta14eccv} integrate depth into the R-CNN approach by Girshick~\etal~\cite{girshick14_rcnn} to detect objects in RGB-D images. They convert depth into 3-channel HHA, \emph{i.e.,} disparity, height and angle encoding and achieve semantic segmentation by training a classifier for superpixels based on the CNN features. Long~\etal~\cite{long15cvpr} propose a fully convolutional network (FCN) which enables end-to-end training for semantic segmentation. Since CNNs reduce the input spatial resolution by a great factor through layers pooling, FCN presents an upsample stage to output high-resolution segmentation by fusing low-resolution predictions. Inspired by FCN and auto-encoders~\cite{yoshua06nips}, encoder-decoder architectures have been proposed to learn upsampling with unpooling and deconvolution~\cite{noh15iccv}.
%In these approaches, the network learns to upsample with successive (de-)convolutions through memorized unpooling~\cite{zeiler11iccv}.
For RGB-D images, Eigen~\etal~\cite{eigen15iccv} propose to train CNNs to predict depth, surface normals and semantics with a multi-task network and achieve very good performance. FuseNet~\cite{lingni16accv} proposes an encoder-decoder CNN to fuse color and depth cues in an end-to-end training for semantic segmentation, which is shown to be more efficient in learning RGB-D features in comparison to direct concatenation of RGB and depth or the use of HHA.
Recently, more complex CNN architectures have been proposed that include multi-resolution refinement~\cite{lin2017_refinenet}, dilated convolutions~\cite{yu2016_dilated} and residual units (e.g.,~\cite{wu2016_widerordeeper}) to achieve state-of-the-art single image semantic segmentation.
Li~\etal~\cite{li2016_lstmcf} use a LSTM recurrent neural network to fuse RGB and depth cues and obtain smooth predictions. Lin~\etal~\cite{lin16exploring} design a CNN that corresponds to a conditional random field (CRF) and use piecewise training to learn both unary and pairwise potentials end-to-end.
%While this method produces very good results, it requires a mean-field approximation for coarse inference and high-resolution refinement using a dense CRF~\cite{krahenb11nips}.
Our approach trains a network on multi-view consistency and fuses the results from multiple view points. It is complementary to the above single-view CNN approaches.

{\bf Semantic SLAM.} In the domain of semantic SLAM, Salas-Moreno~\etal~\cite{moreno14cvpr} developed the SLAM++ algorithm to perform RGB-D tracking and mapping at the object instance level.
%This method works well for indoor scenes which contain many repeated objects with predefined CAD models in a database.
%Tateno~\etal~\cite{tateno15iros} proposed a real-time incremental method to segment RGB-D images based on convex shapes and to smooth the results over the sequence within a SLAM framework.
%Similarily, Zhang~\etal~\cite{zhang15acmgraph} proposed a real-time algorithm for RGB-D mapping, where they classify pixels into plane and non-plane regions and use volumetric fusion to obtained consistent object identity.
%However, the results are pure segments with no additional semantic meaning.
Hermans~\etal~\cite{hermans14icra} proposed 3D semantic mapping for indoor RGB-D sequences based on RGB-D visual odometry and a random forest classifier that performs semantic image segmentation. The individual frame segmentations are projected into 3D and smoothed using a CRF on the point cloud. St\"uckler~\etal~\cite{stueckler15_semslam} perform RGB-D SLAM and probabilistically fuse the semantic segmentations of individual frames obtained with a random forest in multi-resolution voxel maps. Recently, Armeni~\etal~\cite{armeni16cvpr} propose a hierarchical parsing method for large-scale 3D point clouds of indoor environments. They first seperate point clouds into disjoint spaces, \emph{i.e.,} single rooms, and then further cluster points at the object level according to handcrafted features.

{\bf Multi-View Semantic Segmentation.} In contrast to the popularity of CNNs for image-based segmentation, it is less common to apply CNNs for semantic segmentation on multi-view 3D reconstructions.
%This is partially due to the lack of an organized structure in point clouds or the less managable scale of volumetric representations for training a deep neural network.
Recently, Riegler~\etal~\cite{riegler2016_octnet} apply 3D CNNs on sparse octree data structures to perform semantic segmentation on voxels. Nevertheless, the volumetric representations may discard details which are present at the original image resolution. McCormac~\etal~\cite{mccormac2016_semanticfusion} proposed to fuse CNN semantic image segmentations on a 3D surfel map~\cite{whelan16_elasticfusion}.
He~\etal~\cite{he2017_std2p} propose to fuse CNN semantic segmentations from multiple views in video using superpixels and optical flow information.
In contrast to our approach, these methods do not impose multi-view consistency during CNN training and cannot leverage the view-point invariant features learned by our network.
Kundu~\etal~\cite{kundu2016_featurespaceoptim} extend dense CRFs to videos by associating pixels temporally using optical flow and optimizing their feature similarity.
Closely related to our approach for enforcing multi-view consistency is the approach by Su et al.~\cite{su2015_mvcnn} who investigate the task of 3D shape recognition. They render multiple views onto 3D shape models which are fed into a CNN feature extraction stage that is shared across views. The features are max-pooled across view-points and fed into a second CNN stage that is trained for shape recognition. Our approach uses multi-view pooling for the task of semantic segmentation and is trained using realistic imagery and SLAM pose estimates. Our trained network is able to classify single views, but we demonstrate that multi-view fusion using the network trained on multi-view consistency improves segmentation performance over single-view trained networks.

%%%%%%%%%%%%%%%%%%%%%%%%%%%%%%%%%%%%%%%%%%%%%%%%%%%%%%%%%%%%%%%%%%%%%%%%%%%%%%% methods
\section{CNN Architecture for Semantic Segmentation}
In this section, we detail the CNN architecture for semantic segmentation of each RGB-D image of a sequence.
We base our encoder-decoder CNN on FuseNet~\cite{lingni16accv} which learns rich features from RGB-D data.
We enhance the approach with multi-scale loss minimization, which gains additional improvement in segmentation performance.

\subsection{RGB-D Semantic Encoder-Decoder}
Fig.~\ref{fig:network} illustrates our CNN architecture. The network follows an encoder-decoder design, similar to previous work on semantic segmentation~\cite{noh15iccv}. The encoder extracts a hierarchy of features through convolutional layers and aggregates spatial information by pooling layers to increase the receptive field. The encoder outputs low-resolution high-dimensional feature maps, which are upsampled back to the input resolution by the decoder through layers of memorized unpooling and deconvolution. Following FuseNet~\cite{lingni16accv}, the network contains two branches to learn features from RGB ($\mathcal{F}_{rgb}$) and depth ($\mathcal{F}_{d}$), respectively. The feature maps from the depth branch are consistently fused into the RGB branch at each scale. We denote the fusion by~$\mathcal{F}_{rgb}\oplus \mathcal{F}_d$.
% To learn features from RGB-D images, we adopt the FuseNet architecture~\cite{lingni16accv} which is shown to be more efficient in learning features from RGB-D images in comparison to simple concatenation of RGB and depth or to the use of HHA~\cite{gupta14eccv} representation. As demonstrated in Fig.~\ref{fig:network}, the network contains two branches each learning features from RGB ($\mathcal{F}_{rgb}$) and depth ($\mathcal{F}_{d}$), respectively.

The semantic label set is denoted as~$\mathcal{L} = \{1,2,\ldots, K\}$ and the category index is indicated with subscript~$j$. Following notation convention, we compute the classification score~$\mathcal{S}=(s_1, s_2, \ldots , s_K)$ at location~$\mathbf{x}$ and map it to the probability distribution~$\mathcal{P}=(p_1, p_2,\ldots, p_K)$ with the softmax function $\sigma(\cdot)$. Network inference obtains the probability
\begin{equation}\label{eq:softmax}
    p_j(\mathbf{x}, \mathcal{W} \mid \mathcal{I})
    = \sigma(s_j(\mathbf{x}, \mathcal{W})) = \frac{\exp (s_j(\mathbf{x}, \mathcal{W}))}{\sum_k^K \exp (s_k(\mathbf{x}, \mathcal{W}))} \;,
\end{equation}
of all pixels~$\mathbf{x}$ in the image for being labelled as class~$j$, given input RGB-D image~$\mathcal{I}$ and network parameters~$\mathcal{W}$.

We use the cross-entropy loss to learn network parameters for semantic segmentation from ground-truth annotations $l_{gt}$,
\begin{equation}
    L(\mathcal{W}) = - \frac{1}{N}\sum_i^N \sum_j^K \llbracket j =l_{gt} \rrbracket \log p_j(\mathbf{x}_i, \mathcal{W} \mid \mathcal{I})\;,
\end{equation}
where~$N$ is the number of pixels. This loss minimizes the Kullback-Leibler (KL) divergence between predicted distribution and the ground-truth, assuming the ground-truth has a one-hot distribution on the true label.

\begin{figure*}
    \def\imgw{9.2em}
    \centering
    \begin{tikzpicture}[anchor=west,inner sep=0pt]

    \tikzstyle{cmt}=[font=\fontsize]

    \node(label1) at (0,0)    [anchor=west]{\includegraphics[width=\imgw]{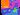}};
    \node(label2) at ([xshift=5pt]label1.east) [anchor=west]{\includegraphics[width=\imgw]{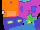}};
    \node(label3) at ([xshift=5pt]label2.east) [anchor=west]{\includegraphics[width=\imgw]{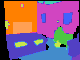}};
    \node(label4) at ([xshift=5pt]label3.east) [anchor=west]{\includegraphics[width=\imgw]{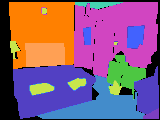}};
    \node(label5) at ([xshift=5pt]label4.east) [anchor=west]{\includegraphics[width=\imgw]{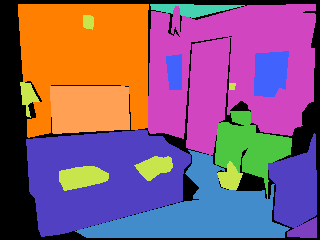}};

    \node(pred1) at (label1.south) [anchor=north, yshift=-1pt]{\includegraphics[width=\imgw]{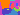}};
    \node(pred2) at ([xshift=5pt]pred1.east) [anchor=west]{\includegraphics[width=\imgw]{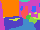}};
    \node(pred3) at ([xshift=5pt]pred2.east) [anchor=west]{\includegraphics[width=\imgw]{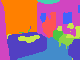}};
    \node(pred4) at ([xshift=5pt]pred3.east) [anchor=west]{\includegraphics[width=\imgw]{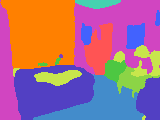}};
    \node(pred5) at ([xshift=5pt]pred4.east) [anchor=west]{\includegraphics[width=\imgw]{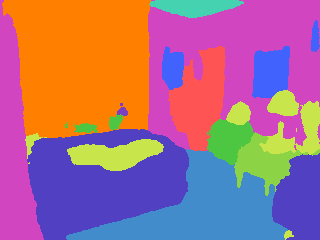}};

    \end{tikzpicture}
    \caption{Example of multi-scale ground-truth and predictions. Upper row: successive subsampled of ground-truth annotation obtained through stochastic pooling. Lower row: CNN prediction on each scale. The resolutions are coarse to fine from left to right with $20\times15$, $40\times30$, $80\times60$, $160\times120$ and $320\times240$.}
    \label{fig:multiscale}
\end{figure*}

% (a) the target $640\times480$ RGB image;
% /*(b) the final $640\times480$ segmentation obtained by upsampling the network output and smooth refine with CRF;*/

\subsection{Multi-Scale Deep Supervision}
The encoder of our network contains five $2\times 2$ pooling layers and downsamples the input resolution by a factor of 32. The decoder learns to refine the low resolution back to the original one with five memorized unpooling followed by deconvolution. In order to guide the decoder through the successive refinement, we adopt the deeply supervised learning method~\cite{lee15aistats, dosovitskiy15iccv} and compute the loss for all upsample scales. For this purpose, we append a classification layer at each deconvolution scale and compute the loss for the respective resolution of ground-truth which is obtained through stochastic pooling~\cite{zeiler2013_stochasticpooling} over the full resolution annotation (see Fig.~\ref{fig:multiscale} for an example).

\section{Multi-View Consistent Learning and Prediction}\label{sec:consistency}
While CNNs have been shown to obtain the state-of-the-art semantic segmentation performances for many datasets, most of these studies focus on single views. When observing a scene from a moving camera such as on a mobile robot, the system obtains multiple different views onto the same objects.
The key innovation of this work is to explore the use of temporal multi-view consistency within RGB-D sequences for CNN training and prediction.
For this purpose, we perform 3D data association by warping multiple frames into a common reference view.
This then enables us to impose multi-view constraints during training.
%We introduce warping layers to impose temporal multi-view consistency which regularizes CNN training and increases prediction accuracy.
In this section, we describe several variants of such constraints.
Notably, these methods can also be used at test time to fuse predictions from multiple views in a reference view.
% We aim to use this for increasing the consistency of semantic maps by fusing semantic image segmentations in keyframes from multiple view points. Moreover, we can make use of the multi-view information in RGB-D video for training a CNN to produce consistent semantic segmentations under view-point changes.

\subsection{Multi-view Data Association Through Warping}
Instead of single-view training, we train our network on RGB-D sequences with poses estimated by a SLAM algorithm. We define each training sequence to contain one reference view~$\mathcal{I}_k$ with ground-truth semantic annotations and several overlapping views~$\mathcal{I}_i$ that are tracked towards $\mathcal{I}_k$. The relative poses~$\boldsymbol{\xi}$ of the neighboring frames are estimated through tracking algorithms such as DVO SLAM~\cite{kerl13iros}.
In order to impose temporal consistency, we adopt the warping concept from multi-view geometry to associate pixels between view points and introduce warping layers into our CNN. The warping layers synthesize CNN output in a reference view from a different view at any resolution by sampling given a known pose estimate and the known depth. The warping layers can be viewed as a variant of spatial transformers~\cite{jaderberg15nips} with fixed transformation parameters.

We now formulate the warping. Given 2D image coordinate $\mathbf{x}\in\mathbb{R}^2$, the warped pixel location
\begin{equation}\label{eq:warping}
    \mathbf{x}^\omega := \omega(\mathbf{x}, \boldsymbol{\xi})=\pi \big( \mathbf{T}(\boldsymbol{\xi}) \, \pi^{-1}(\mathbf{x}, Z_i(\mathbf{x}))\big) \;,
\end{equation}
is determined through the warping function~$\omega(\mathbf{x}, \boldsymbol{\xi})$ which transforms the location from one camera view to the other using the depth~$Z_i(\mathbf{x})$ at pixel~$\mathbf{x}$ in image~$\mathcal{I}_i$ and the SLAM pose estimate~$\boldsymbol{\xi}$. The functions~$\pi$ and its inverse~$\pi^{-1}$ project homogeneous 3D coordinates to image coordinates and vice versa, while~$\mathbf{T}(\boldsymbol{\xi})$ denotes the homogeneous transformation matrix derived from pose~$\boldsymbol{\xi}$.

Using this association by warping, we synthesize the output of the reference view by sampling the feature maps of neighboring views using bilinear interpolation. Since the interpolation is differentiable, it is straight-forward to back-propagate gradients through the warping layers. With a slight abuse of notation, we denote the operation of synthesizing the layer output~$\mathcal{F}$ given the warping by $\mathcal{F}^\omega := \mathcal{F}(\omega( \mathbf{x}, \boldsymbol{\xi}))$.
%In Section~\ref{sec:consistency}, we further discuss how the warping layers facilitate multi-view consistency.

We also apply deep supervision when training for multi-view consistency through warping. As shown in Fig.~\ref{fig:network}, feature maps at each resolution of the decoder are warped into the common reference view. Despite the need to perform warping at multiple scales, the warping grid is only required to be computed once at the input resolution, and is normalized to the canonical coordinates within the range of $[-1, 1]$. The lower-resolution warping grids can then be efficiently generated through average pooling layers.

\subsection{Consistency Through Warp Augmentation}
One straight-forward solution to enforce multi-view segmentation consistency is to warp the predictions of neighboring frames into the ground-truth annotated keyframe and computing a supervised loss there. This approach can be interpreted as a type of data augmentation using the available nearby frames. We implement this consistency method by warping the keyframe into neighboring frames, and synthesize the classification score of the nearby frame from the keyframe's view point. We then compute the cross-entropy loss on this synthesized prediction. Within RGB-D sequences, objects can appear at various scales, image locations, view perspective, color distortion given uncontrolled lighting and shape distortion given rolling shutters of RGB-D cameras. Propagating the keyframe annotation into other frames implicitly regulates the network predictions to be invariant under these transformations.

\subsection{Consistency Through Bayesian Fusion}

Given a sequence of measurements and predictions at test time, Bayesian fusion is frequently applied to aggregate the semantic segmentations of individual views. Let us denote the semantic labelling of a pixel by~$y$ and its measurement in frame~$i$ by~$z_i$. We use the notation~$z^i$ for the set of measurements up to frame~$i$. According to Bayes rule,
\begin{align}\label{eq:b1}
    p( y \mid z^i ) &= \frac{p( z_i \mid y, z^{i-1} ) \, p( y \mid z^{i-1} )}{p( z_i \mid z^{i-1} )} \\
    &= \eta_i \, p( z_i \mid y, z^{i-1} ) \, p( y \mid z^{i-1} ) \;.
\end{align}
Suppose measurements satisfy the \emph{i.i.d.} condition, i.e. $p( z_i \mid y, z^{i-1} ) = p( z_i \mid y )$, and equal a-priori probability for each class, then Equation~\eqref{eq:b1} simplifies to
\begin{equation}\label{eq:b2}
    p( y \mid z^i ) = \eta_i \, p( z_i \mid y ) \, p( y \mid z^{i-1} ) = \prod_{i} \eta_i \, p( z_i \mid y ) \;.
\end{equation}
Put simple, Bayesian fusion can be implemented by taking the product over the semantic labelling likelihoods of individual frame at a pixel and normalizing the product to yield a valid probability distribution. This process can also be implemented recursively on a sequence of frames.

When training our CNN for multi-view consistency using Bayesian fusion, we warp the predictions of neighboring frames into the keyframe using the SLAM pose estimate. We obtain the fused prediction at each keyframe pixel by summing up the unnormalized log labelling likelihoods instead of the individual frame softmax outputs. Applying softmax on the sum of log labelling likelihoods yields the fused labelling distribution. This is equivalent to Eq.~\eqref{eq:b2} since
\begin{equation}\label{eq:fusioneq}
    \frac{\prod_{i} p_{i,j}^\omega}{\sum_k^K \prod_i p_{i,k}^\omega}
    = \frac{\prod_{i} \sigma(s_{i,j}^\omega)} {\sum_k^K \prod_i \sigma(s_{i,k}^\omega)}
    = \sigma\left( \sum_i s_{i,j}^\omega \right) \;,
\end{equation}
where $s_{i,j}^\omega$ and $p_{i,j}^\omega$ denote the warped classification scores and probabilities, respectively, and $\sigma(\cdot)$ is the softmax function as defined in Equation~\eqref{eq:softmax}.

\subsection{Consistency Through Multi-View Max-Pooling}

While Bayesian fusion provides an approach to integrate several measurements in the probability space, we also explore direct fusion in the feature space using multi-view max-pooling of the warped feature maps. We warp the feature maps preceeding the classification layers at each scale in our decoder into the keyframe and apply max-pooling over corresponding feature activations at the same warped location to obtain a pooled feature map in the keyframe,
\begin{equation}
    \mathcal{F} = \operatorname{max\_pool} (\mathcal{F}^\omega_1, \mathcal{F}^\omega_2, \ldots, \mathcal{F}^\omega_N) \;.
\end{equation}
The fused feature maps are classified and the resulting semantic segmentation is compared to the keyframe ground-truth for loss calculation.

%%%%%%%%%%%%%%%%%%%%%%%%%%%%%%%%%%%%%%%%%%%%%%%%%%%%%%%%%%%%%%%%%%%%%%%%%%%%%%% evaluation
\section{Evaluation}
We evaluate our proposed approach using the NYUDv2 RGB-D dataset~\cite{silberman12eccv}. The dataset provides 1449 pixelwise annotated RGB-D images capturing various indoor scenes, and is split into 795 frames for training/validation (trainval) and 654 frames for testing. The original sequences that contain these 1449 images are also available with NYUDv2, whereas sequences are unfortunately not available for other large RGB-D semantic segmentation datasets. Using DVO-SLAM~\cite{kerl13iros}, we determine the camera poses of neighboring frames around each annotated keyframe to obtain multi-view sequences. This provides us with in total 267,675 RGB-D images, despite that tracking fails for 30 out of 1449 keyframes. Following the original trainval/test split, we use 770 sequences with 143,670 frames for training and 649 sequences with 124,005 frames for testing. For benchmarking, our method is evaluated for the 13-class~\cite{couprie13iclr} and 40-class~\cite{gupta13cvpr} semantic segmentation tasks. We use the raw depth images without inpainted missing values.

\subsection{Training Details}
We implemented our approach using the Caffe framework~\cite{jia14caffe}. For all experiments, the network parameters are initialized as follows. The convolutional kernels in the encoder are initialized with the pretrained 16-layer VGGNet~\cite{simonyan14vgg} and the deconvolutional kernels in the decoder are initialized using He's method~\cite{he15iccv}. For the first layer of the depth encoder, we average the original three-channel VGG weights to obtain a single-channel kernel. We train the network with stochastic gradient descent (SGD)~\cite{bottou12sgd} with 0.9 momentum and 0.0005 weight decay. The learning rate is set to 0.001 and decays by a factor of 0.9 every 30k iterations. All the images are resized to a resolution of $320\times240$ pixels as input to the network and the predictions are also up to this scale. To downsample, we use cubic interpolation for RGB images and nearest-neighbor interpolation for depth and label images. During training, we use a minibatch of 6 that comprises two sequences, with one keyframe and two tracking frames for each sequence. We apply random shuffling after each epoch for both inter and intra sequences. The network is trained until convergence. We observed that multi-view CNN training does not require significant extra iterations for convergence. For multi-view training, we sample from the nearest frames first and include 10 further-away frames every 5 epochs. By this, we alleviate that typically tracking errors accumulate and image overlap decreases as the camera moves away from the keyframe.

\subsection{Evaluation Criteria}
We measure the semantic segmentation performance with three criteria: global pixelwise accuracy, average classwise accuracy and average intersection-over-union (IoU) scores. These three criteria can be calculated from the confusion matrix. With $K$ classes, each entry of the $K\times K$ confusion matrix~$c_{ij}$ is the total amount of pixels belonging to class~$i$ that are predicted to be class~$j$. The global pixelwise accuracy is computed by $\sum_{i}c_{ii} / \sum_{ij}c_{ij}$, the average classwise accuracy is computed by $\frac{1}{K} \sum_{i} (c_{ii} / \sum_j c_{ij})$, and the average IoU score is calculated by $\frac{1}{K} \sum_i \big(c_{ii} / (\sum_i c_{ij} + \sum_j c_{ij} - c_{ii}) \big)$.

\begin{table}[t!]
    \centering
    \caption{Single-view semantic segmentation accuracy of our network in comparison to the state-of-the-art methods for NYUDv2 13-class and 40-class segmentation tasks.}
    \label{tab:monotest}
    \begin{tabular}{C{4ex} L{20ex} L{12ex} C{9ex} C{9ex} C{4ex} }
    \toprule
    &methods                                 &input     &pixelwise  &classwise &IoU\\

    % \cmidrule{1-6}
    \midrule
    \multirow{9}{*}{\rot{\shortstack[c]{NYUDv2 \\ 13 classes}}}
    &Couprie~\etal~\cite{couprie13iclr}         &RGB-D          & 52.4 & 36.2 &-\\    %% RGBD multiscale, from their paper
    &Hermans~\etal~\cite{hermans14icra}         &RGB-D          & 54.2 & 48.0 &-\\     %% RGBD + crf from their paper
    &SceneNet~\cite{ankur16cvpr}                &DHA            & 67.2 & 52.5 &-\\    %% DHA from their paper
    &Eigen~\etal~\cite{eigen15iccv}             &RGB-D-N        & 75.4 & 66.9 & 52.6 \\    %% RGBD + normal. VGG based model from their paper, i tested with a different accuracy
    &FuseNet-SF3~\cite{lingni16accv}            &RGB-D          & 75.8 & 66.2 & 54.2 \\ % 190k with data augmentation, \fixme{show the numbers without augmentation}
    % \cmidrule{2-6}
    &MVCNet-Mono                                 &RGB-D          & 77.6 & 68.7 & 56.9 \\ % 290k, b6 with data augmentation
    &MVCNet-Augment                              &RGB-D          & 77.6 & 69.3 & 57.2 \\ % 175k
    &MVCNet-Bayesian                             &RGB-D          & \bf 77.8 & \it 69.4 & \bf 57.3 \\ % 105k
    &MVCNet-MaxPool                              &RGB-D          & \it 77.7 & \bf 69.5 & \bf 57.3 \\ % 80k
    \cmidrule{2-6}

    \multirow{9}{*}{\rot{\shortstack[c]{NYUDv2 \\40 classes}}}
    &RCNN~\cite{gupta14eccv}                    &RGB-HHA        & 60.3 & 35.1 & 28.6\\     %% RGB-HHA from their paper
    &FCN-16s~\cite{long15cvpr}                  &RGB-HHA        & 65.4 & 46.1 & 34.0\\  %% rgb-hha from their paper
    &Eigen~\etal~\cite{eigen15iccv}             &RGB-D-N        & 65.6 & 45.1 & 34.1\\  %% take from their paper
    &FuseNet-SF3~\cite{lingni16accv}            &RGB-D          & 66.4 & 44.2 & 34.0\\  %% 230k
    &Context-CRF~\cite{lin16exploring}          &RGB            & 67.6 & 49.6 & 37.1\\  %%
    % \cmidrule{2-6}
    &MVCNet-Mono                                 &RGB-D          & \it 68.6 & 48.7 & 37.6\\
    &MVCNet-Augment                              &RGB-D          & \it 68.6 & \it 49.9 & \bf 38.0\\ %%210k
    &MVCNet-Bayesian                             &RGB-D          & 68.4 & 49.5 & 37.4\\
    &MVCNet-MaxPool                              &RGB-D          & \bf 69.1 & \bf 50.1 & \bf 38.0\\ %%40
    \bottomrule
    \end{tabular}
\end{table}

\begin{table}[t!]
    \centering
    \caption{Multi-view segmentation accuracy of our network using Bayesian fusion for NYUDv2 13-class and 40-class segmentation.}
    \label{tab:fusiontest}
    \begin{tabular}{C{4ex} L{20ex} C{11ex} C{11ex} C{11ex} }
    \toprule
    &methods                                &pixelwise  &classwise &IoU\\
    % \cmidrule{1-5}
    \midrule
    \multirow{5}{*}{\rot{\shortstack[c]{NYUDv2 \\ 13 classes}}}
    &FuseNet-SF3~\cite{lingni16accv}        &77.19 &67.46 &56.01 \\  % 190k with data augmentation
    &MVCNet-Mono                             &78.70 &69.61 &58.29 \\  % 290k with data augmentation
    &MVCNet-Augment                          &78.94 &\it 70.48 &58.93 \\  % 175k
    &MVCNet-Bayesian                         &\bf 79.13 &\it 70.48 &\it 59.04 \\  % 105
    &MVCNet-MaxPool                         &\bf 79.13 &\bf 70.59 &\bf 59.07 \\  % 80
    \cmidrule{2-5}
    \multirow{5}{*}{\rot{\shortstack[c]{NYUDv2 \\40 classes}}}
    &FuseNet-SF3~\cite{lingni16accv}        &67.74 & 44.92 & 35.36 \\ %
    &MVCNet-Mono                             &70.03 & 49.73 & 39.12 \\ % 290k with data augmentation
    &MVCNet-Augment                          &\it 70.34 & \it 51.73 & \bf 40.19\\ % 210k
    &MVCNet-Bayesian                         &70.24 & 51.18 & 39.74\\
    &MVCNet-MaxPool                         &\bf 70.66 & \bf 51.78 & \it 40.07\\ % 40k
    \bottomrule
    \end{tabular}
\end{table}

\begin{table*}
    \centering
    \caption{NYUDv2 13-class semantic segmentation IoU scores. Our method achieves best per-class accuracy and average IoU.}%  In particular, we obtain the best IoU for 9 out of 13 classes. The same color code is used in Figure~\ref{fig:visualcmp}.}
    \label{tab:classwise}
    \setlength{\tabcolsep}{4.5pt}
    \begin{tabular}{C{4ex} L{24ex}  cl*{16}{r} }
        \toprule
        &method
        &\block{icra1}{bed} &\block{icra2}{objects} &\block{icra3}{chair} &\block{icra4}{furniture} &\block{icra5}{ceiling} &\block{icra6}{floor} &\block{icra7}{decorat.}
        &\block{icra8}{sofa} &\block{icra9}{table} &\block{icra10}{wall} &\block{icra11}{window} &\block{icra12}{books} &\block{icra13}{TV}
        &\rot{\shortstack[l]{average\\accuracy}}\\
        % \cmidrule{2-16}
        \midrule
        &class frequency                 &4.08  &7.31  &3.45  &12.71 &1.47  &9.88  &3.40  &2.84  &3.42  &24.57 &4.91  &2.78  &0.99 & \\
        % \cmidrule{2-16}
        \midrule

        \multirow{6}{*}{\rot{\shortstack[c]{single-view}}}
        &Eigen~\etal~\cite{eigen15iccv}  &56.71 &38.29 &50.23 &54.76 &64.50 &89.76 &45.20 &47.85 &42.47 &74.34 &56.24 &45.72 &34.34 &53.88\\ %%
        &FuseNet-SF3~\cite{lingni16accv} &61.52 &37.95 &52.67 &53.97 &64.73 &89.01 &47.11 &57.17 &39.20 &75.08 &58.06 &37.64 &29.77 &54.14\\
        &MVCNet-Mono                      &65.27 &37.82 &54.09 &59.39 &65.26 &89.15 &49.47 &57.00 &44.14 &75.31 &57.22 &49.21 &36.14 &56.88\\ %290
        &MVCNet-Augment                   &65.33 &38.30 &54.15 &\bf59.54 &\bf67.65 &89.26 &49.27 &55.18 &43.39 &74.59 &58.46 &\bf49.35 &\bf38.84 &57.18\\ %175
        &MVCNet-Bayesian                  &\bf65.76 &38.79 &\bf54.60 &59.28 &67.58 &89.69 &48.98 &\bf56.72 &42.42 &75.26 &\bf59.55 &49.27 &36.51 &57.26\\
        &MVCNet-MaxPool                  &65.71 &\bf39.10 &54.59 &59.23 &66.41 &\bf89.94 &\bf49.50 &56.30 &\bf43.51 &\bf75.33 &59.11 &49.18 &37.37 &\bf57.33\\

        \cmidrule{2-16}
        \multirow{5}{*}{\rot{\shortstack[c]{multi-view}}}
        &FuseNet-SF3~\cite{lingni16accv} &64.95 &39.62 &55.28 &55.90 &64.99 &89.88 &47.99 &\bf60.17 &42.40 &76.24 &59.97 &39.80 &30.91 &56.01\\ %% 190k,
        &MVCNet-Mono                      &67.11 &40.14 &56.39 &60.90 &66.07 &89.77 &50.32 &59.49 &46.12 &76.51 &59.03 &48.80 &37.13 &58.29\\
        &MVCNet-Augment                   &68.22 &40.04 &56.55 &61.82 &67.88 &90.06 &50.85 &58.00 &\bf45.98 &75.85 &60.43 &50.50 &\bf39.89 &58.93\\
        &MVCNet-Bayesian                  &\bf68.38 &\bf40.87 &\bf57.10 &\bf61.84 &\bf67.98 &\bf90.64 &50.05 &59.70 &44.73 &76.50 &\bf61.75 &\bf51.01 &36.99 &59.04\\
        &MVCNet-MaxPool                  &68.09 &41.58 &56.88 &61.56 &67.21 &\bf90.64 &50.69 &59.73 &45.46 &\bf76.68 &61.28 &50.60 &37.51 &\bf59.07\\
        \bottomrule
    \end{tabular}
\end{table*}
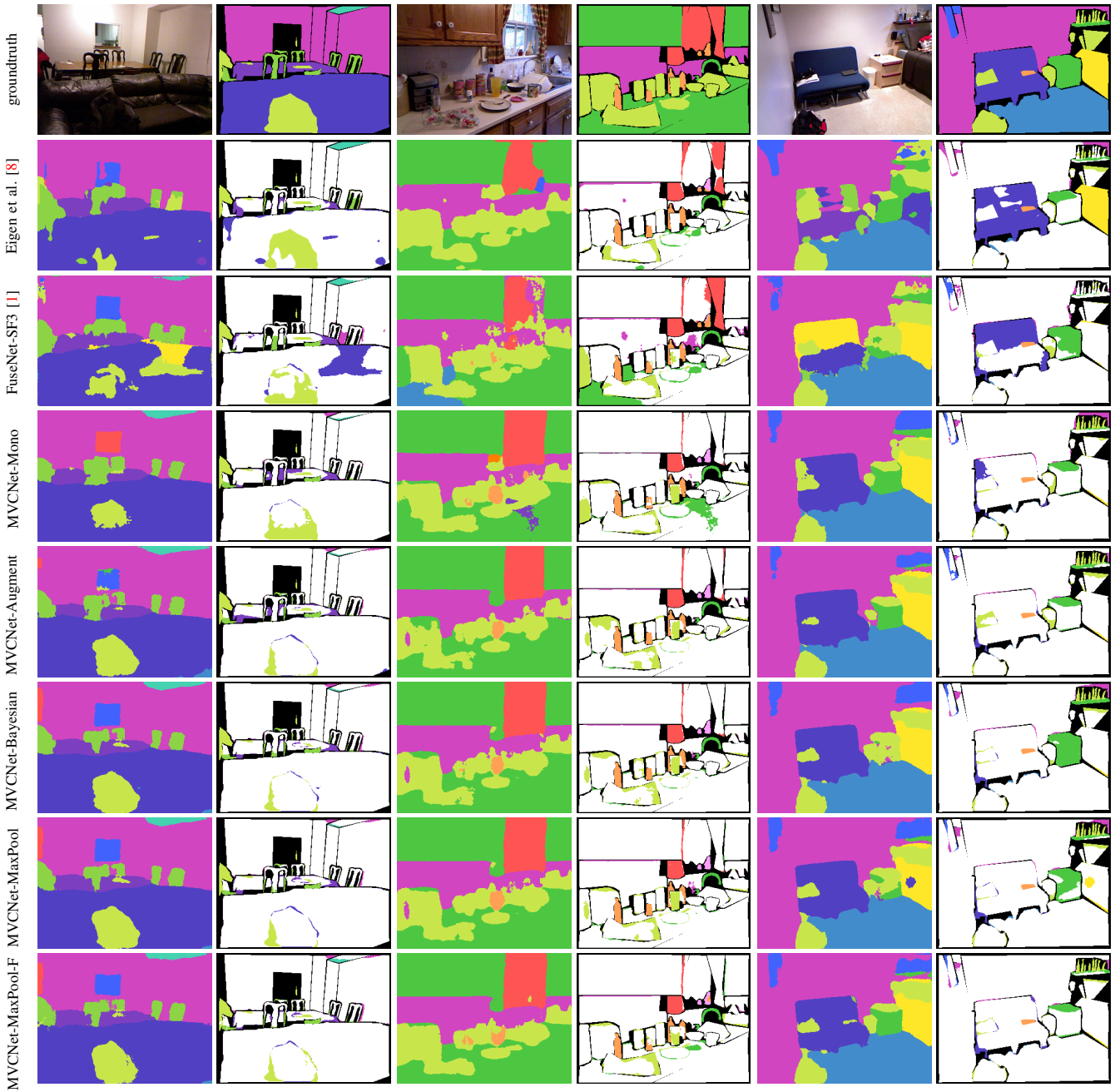
\begin{figure*}
\centering
\begin{tikzpicture}[inner sep=1pt]
    \def\imwidth{7.9em}
    \def\imh{6.2em}
    \def\imsep{2mm}
    \tikzstyle{cmt}=[rotate=90,anchor=center, font=\footnotesize, text centered]

    \def\onecmp#1#2
    {
    \begin{scope}
        \node(rgb)    at(#2,0)[anchor=west] {\includegraphics[width=\imwidth]{rgb#1.png}};
        \node(gt)     at(rgb.south)[anchor=north] {\includegraphics[width=\imwidth]{gt#1.png}};
        \node(eigen)  at(gt.south)[anchor=north] {\includegraphics[width=\imwidth]{eigen#1.png}};
        \node(sf)     at(eigen.south)[anchor=north]{\includegraphics[width=\imwidth]{sf3#1.png}};
        \node(mono)   at(sf.south)[anchor=north] {\includegraphics[width=\imwidth]{mono#1.png}};
        \node(aug)    at(mono.south)[anchor=north] {\includegraphics[width=\imwidth]{aug#1.png}};
        \node(baye)   at(aug.south)[anchor=north] {\includegraphics[width=\imwidth]{bayesian#1.png}};
        \node(mvp)    at(baye.south)[anchor=north] {\includegraphics[width=\imwidth]{mvp#1.png}};
        \node(mvpf)   at(mvp.south)[anchor=north] {\includegraphics[width=\imwidth]{mvp-fuse#1.png}};
    \end{scope}
    }
    % \onecmp{210}{0}
    % \onecmp{551}{\imwidth  +\imsep}
    % \onecmp{280}{\imwidth*2+\imsep*2}
    % \onecmp{561}{\imwidth*3+\imsep*3} %300
    % \onecmp{207}{\imwidth*4+\imsep*4} %934

    \def\doublecmp#1#2
    {
    \begin{scope}
        \node(rgb)    at(#2,0)[anchor=west] {\includegraphics[width=\imwidth]{rgb#1.png}};
        \node(eigen)  at(rgb.south)[anchor=north] {\includegraphics[width=\imwidth]{eigen#1.png}};
        \node(sf)     at(eigen.south)[anchor=north]{\includegraphics[width=\imwidth]{sf3#1.png}};
        \node(mono)   at(sf.south)[anchor=north] {\includegraphics[width=\imwidth]{mono#1.png}};
        \node(aug)    at(mono.south)[anchor=north] {\includegraphics[width=\imwidth]{aug#1.png}};
        \node(baye)   at(aug.south)[anchor=north] {\includegraphics[width=\imwidth]{bayesian#1.png}};
        \node(mvp)    at(baye.south)[anchor=north] {\includegraphics[width=\imwidth]{mvp#1.png}};
        \node(mvpf)   at(mvp.south)[anchor=north] {\includegraphics[width=\imwidth]{mvp-fuse#1.png}};

        \node(gt)     at(rgb.east)[anchor=west] {\includegraphics[width=\imwidth]{gt#1.png}};
        \node(eigen2) at(eigen.east)[anchor=west] {\includegraphics[width=\imwidth]{residual-eigen#1.png}};
        \node(sf2)    at(sf.east)[anchor=west]{\includegraphics[width=\imwidth]{residual-sf3#1.png}};
        \node(mono2)  at(mono.east)[anchor=west] {\includegraphics[width=\imwidth]{residual-mono#1.png}};
        \node(aug2)   at(aug.east)[anchor=west] {\includegraphics[width=\imwidth]{residual-aug#1.png}};
        \node(baye2)  at(baye.east)[anchor=west] {\includegraphics[width=\imwidth]{residual-bayesian#1.png}};
        \node(mvp2)   at(mvp.east)[anchor=west] {\includegraphics[width=\imwidth]{residual-mvp#1.png}};
        \node(mvpf2)  at(mvpf.east)[anchor=west] {\includegraphics[width=\imwidth]{residual-mvp-fuse#1.png}};

    \end{scope}
    }

    \doublecmp{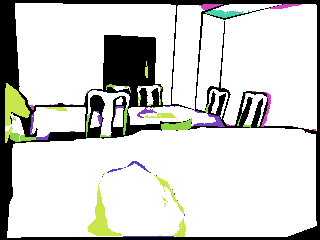}{0}
    \doublecmp{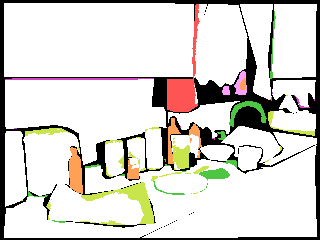}{\imwidth*2  +\imsep}
    \doublecmp{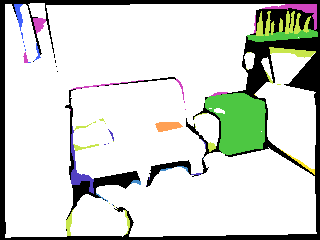}{\imwidth*4+\imsep*2}
    % \doublecmp{561}{\imwidth*3+\imsep*3} %300
    % \doublecmp{207}{\imwidth*4+\imsep*4} %934

    \begin{scope}[xshift=-1em]

    \node(t1) at(0,0) [cmt,] {\scriptsize groundtruth};
    % \node(t2) at(0,-\imh) [cmt, ] {groundtruth};
    \node(t3) at(0,-\imh*1) [cmt,] {\scriptsize Eigen~\etal~\cite{eigen15iccv}};
    \node(t4) at(0,-\imh*2)[cmt,] {\scriptsize FuseNet-SF3~\cite{lingni16accv}};
    \node(t5) at(0,-\imh*3)[cmt,] {\scriptsize MVCNet-Mono};
    \node(t6) at(0,-\imh*4)[cmt,] {\scriptsize MVCNet-Augment};
    \node(t7) at(0,-\imh*5)[cmt,] {\scriptsize MVCNet-Bayesian};
    \node(t8) at(0,-\imh*6)[cmt,] {\scriptsize MVCNet-MaxPool};
    \node(t9) at(0,-\imh*7)[cmt,] {\scriptsize MVCNet-MaxPool-F};
    \end{scope}

\end{tikzpicture}
\caption{Qualitative semantic segmentation results of our methods and several state-of-the-art baselines on NYUDv2 13-class segmentation (see Table~\ref{tab:classwise} for color coding, left columns: semantic segmentation, right columns: falsely classified pixels, black is void). Our multi-view consistency trained models produce more accurate and homogeneous results than single-view methods. Bayesian fusion further improves segmentation quality (e.g. MVCNet-MaxPool-F).}
\label{fig:visualcmp}
\end{figure*}

\begin{figure}
\centering
\begin{tikzpicture}[inner sep=1pt]
    \def\imwidth{8em}
    \def\imh{7.35em}
    \def\imsep{0.1em}
    \tikzstyle{cmt}=[rotate=90,anchor=center, font=\footnotesize, text centered]

    \def\onecmp#1#2
    {
    \begin{scope}
        \node(rgb)    at(#2,0)[anchor=west] {\includegraphics[width=\imwidth]{rgb#1.png}};
        \node(gt)     at(rgb.south)[anchor=north] {\includegraphics[width=\imwidth]{gt#1.png}};
        \node(pred)   at(gt.south)[anchor=north] {\includegraphics[width=\imwidth]{mvp#1.png}};
        \node(pred2)  at(pred.south)[anchor=north] {\includegraphics[width=\imwidth]{residual-mvp#1.png}};
        \node(predf)  at(pred2.south)[anchor=north] {\includegraphics[width=\imwidth]{mvp-fuse#1.png}};
        \node(predf2) at(predf.south)[anchor=north] {\includegraphics[width=\imwidth]{residual-mvp-fuse#1.png}};
    \end{scope}
    }
    \onecmp{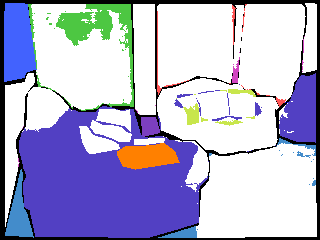}{0}
    \onecmp{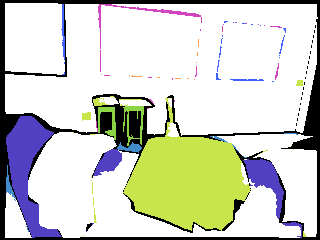}{\imwidth  +\imsep}
    \onecmp{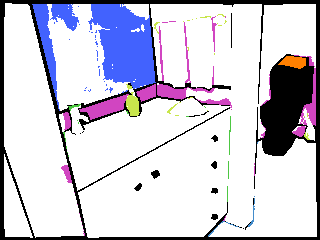}{\imwidth*2+\imsep*2}
\end{tikzpicture}

\caption{Challenging cases for MVCNet-MaxPool-F (top to bottom: RGB image, ground-truth, single-view prediction on keyframe, multi-view prediction fused in keyframe). On the left, the network fails to classify the objects for all frames. In the middle, the network makes some errors in single-view prediction, but through multi-view fusion, some mistakes are corrected. On the right, multi-view fusion degenerates performance due to the mirror reflections. }
\label{fig:bageg}
\end{figure}

\subsection{Single Frame Segmentation}

In a first set of experiments, we evaluate the performance of several variants of our network for direct semantic segmentation of individual frames. This means we do not fuse predictions from nearby frames to obtain the final prediction in a frame. We predict semantic segmentation with our trained models on the 654 test images of the NYUDv2 dataset and compare our methods with state-of-art approaches. The results are shown in Table~\ref{tab:monotest}. Unless otherwise stated, we take the results from the original papers for comparison and report their best results (i.e. SceneNet-FT-NYU-DO-DHA model for SceneNet~\cite{ankur16cvpr}, VGG-based model for Eigen~\etal~\cite{eigen15iccv}). The result of Hermans~\etal~\cite{hermans14icra} is obtained after applying a dense CRF~\cite{krahenb11nips} for each image and in-between neighboring 3D points to further smoothen their results. We also remark that the results reported here for the Context-CRF model are finetuned on NYUDv2 like in our approach to facilitate comparison. Furthermore, the network output is refined using a dense CRF~\cite{krahenb11nips} which is claimed to increase the accuracy of the network by approximately 2\%. The results for FuseNet-SF3 are obtained by our own implementation. Our baseline model MVCNet-Mono is trained without multi-view consistency, which amounts to FuseNet with multiscale deeply supervised loss at decoder. However, we apply single image augmentation to train the FuseNet-SF3 and MVCNet-Mono with random scaling between $[0.8,1.2]$, random crop and mirror. This data augmentation is not used fro multi-view training. Nevertherless, our results show that the different variants of multi-view consistency training outperform the state-of-art methods for single image semantic segmentation. Overall, multi-view max-pooling (MVCNet-MaxPool) has a small advantage over the other multi-view consistency training approaches (MVCNet-Augment and MVCNet-Bayesian).

\subsection{Multi-View Fused Segmentation}

Since we train on sequences, in the second set of experiment, we also evaluate the fused semantic segmentation over the test sequences. The number of fused frames is fixed to 50, which are uniformly sampled over the entire sequence. Due to the lack of ground-truth for neighboring frames, we fuse the prediction of neighboring frames in the keyframes using Bayesian fusion according to Equation~\eqref{eq:fusioneq}. This fusion is typically applied for semantic mapping using RGB-D SLAM. The results are shown in Table~\ref{tab:fusiontest}. Bayesian multi-view fusion improves the semantic segmentation by approx. 2\% on all evaluation measures towards single-view segmentation. Also, the training for multi-view consistency achieves a stronger gain over single-view training (MVCNet-Mono) when fusing segmentations compared to single-view segmentation.
This performance gain is observed in the qualitative results in Fig.~\ref{fig:visualcmp}. It can be seen that our multi-view consistency training and Bayesian fusion produces more accurate and homogeneous segmentations. Fig.~\ref{fig:bageg} shows typical challenging cases for our model.

We also compare classwise and average IoU scores for 13-class semantic segmentation on NYUDv2 in Table~\ref{tab:classwise}.
The results of Eigen~\etal~\cite{eigen15iccv} are from their publicly available model
%\footnote{Available at:~\url{http://cs.nyu.edu/~deigen/dnl/}}
tested on $320\times240$ resolution.
The results demonstrate that our approach gives high performance gains across all occurence frequencies of the classes in the dataset.

% Last, we also show the final semantic maps obtained from our complete pipeline in Figure~\ref{fig:finalmap}, with the fusion from consecutive frames into the keyframe and the dense 3D CRF to smooth the prediction.
% \input{./tikz/plot_framefusion.tex}

%\addtolength{\textheight}{-0.25cm}   % This command serves to balance the column lengths

%%%%%%%%%%%%%%%%%%%%%%%%%%%%%%%%%%%%%%%%%%%%%%%%%%%%%%%%%%%%%%%%%%%%%%%%%%%%%%%%
\section{Conclusion}

In this paper we propose methods for enforcing multi-view consistency during the training of CNN models for semantic RGB-D image segmentation. We base our CNN design on FuseNet~\cite{lingni16accv}, a recently proposed CNN architecture in an encoder-decoder scheme for semantic segmentation of RGB-D images. We augment the network with multi-scale loss supervision to improve its performance. We present and evaluate three different approaches for multi-view consistency training. Our methods use an RGB-D SLAM trajectory estimate to warp semantic segmentations or feature maps from one view point to another. Multi-view max-pooling of feature maps overall provides the best performance gains in single-view segmentation and fusion of multiple views.

We demonstrate the superior performance of multi-view consistency training and Bayesian fusion on the NYUDv2 13-class and 40-class semantic segmentation benchmark. All multi-view consistency training approaches outperform single-view trained baselines. They are key to boosting segmentation performance when fusing network predictions from multiple view points during testing. On NYUDv2, our model sets a new state-of-the-art performance using an end-to-end trained network for single-view predictions as well as multi-view fused semantic segmentation without further postprocessing stages such as dense CRFs. In future work, we want to further investigate integration of our approach in a semantic SLAM system, for example, through coupling of pose tracking and SLAM with our semantic predictions.

\balance

\bibliographystyle{ieeetr}
\bibliography{ms}

\begin{thebibliography}{10}

\bibitem{lingni16accv}
C.~Hazirbas, L.~Ma, C.~Domokos, and D.~Cremers, ``Fusenet: incorporating depth
  into semantic segmentation via fusion-based cnn architecture,'' in {\em Asian
  Conf. on Computer Vision (ACCV)}, 2016.

\bibitem{couprie13iclr}
C.~Couprie, C.~Farabet, L.~Najman, and Y.~Lecun, {\em Indoor semantic
  segmentation using depth information}.
\newblock 2013.

\bibitem{gupta14eccv}
S.~Gupta, R.~Girshick, P.~Arbel{\'a}ez, and J.~Malik, ``Learning rich features
  from {RGB-D} images for object detection and segmentation,'' in {\em Europ.
  Conf. on Computer Vision (ECCV)}, 2014.

\bibitem{girshick14_rcnn}
R.~Girshick, J.~Donahue, T.~Darrell, and J.~Malik, ``Rich feature hierarchies
  for accurate object detection and semantic segmentation,'' in {\em IEEE
  Computer Vision and Pattern Recognition (CVPR)}, 2014.

\bibitem{long15cvpr}
J.~Long, E.~Shelhamer, and T.~Darrell, ``Fully convolutional networks for
  semantic segmentation,'' in {\em IEEE Int. Conf. on Computer Vision and
  Pattern Recognition (CVPR)}, 2015.

\bibitem{yoshua06nips}
Y.~Bengio, P.~Lamblin, D.~Popovici, and H.~Larochelle, ``Greedy layer-wise
  training of deep networks,'' in {\em Advances in Neural Information
  Processing Systems (NIPS)}, 2007.

\bibitem{noh15iccv}
H.~Noh, S.~Hong, and B.~Han, ``Learning deconvolution network for semantic
  segmentation,'' in {\em IEEE Int. Conf. on Computer Vision and Pattern
  Recognition (CVPR)}, pp.~1520--1528, 2015.

\bibitem{eigen15iccv}
D.~Eigen and R.~Fergus, ``Predicting depth, surface normals and semantic labels
  with a common multi-scale convolutional architecture,'' in {\em IEEE Int.
  Conf. on Computer Vision (ICCV)}, 2015.

\bibitem{lin2017_refinenet}
G.~Lin, A.~Milan, C.~Shen, and I.~Reid, ``Refine{N}et: {M}ulti-path refinement
  networks for high-resolution semantic segmentation,'' in {\em IEEE Computer
  Vision and Pattern Recognition (CVPR)}, 2017.

\bibitem{yu2016_dilated}
F.~Yu and V.~Koltun, ``Multi-scale context aggregation by dilated
  convolutions,'' in {\em Int. Conf. on Learning Representations (ICLR)}, 2016.

\bibitem{wu2016_widerordeeper}
Z.~Wu, C.~Shen, and A.~van~den Hengel, ``Wider or deeper: Revisiting the resnet
  model for visual recognition,'' {\em CoRR abs/1611.10080}, 2016.

\bibitem{li2016_lstmcf}
Z.~Li, Y.~Gan, X.~Liang, Y.~Yu, H.~Cheng, and L.~Lin, ``{LSTM-CF}: Unifying
  context modeling and fusion with {LSTMs} for {RGB-D} scene labeling,'' in
  {\em Europ. Conf. on Computer Vision (ECCV)}, 2016.

\bibitem{lin16exploring}
G.~Lin, C.~Shen, A.~van~den Hengel, and I.~D. Reid, ``Exploring context with
  deep structured models for semantic segmentation,'' {\em CoRR},
  vol.~abs/1603.03183, 2016.

\bibitem{moreno14cvpr}
R.~F. Salas-Moreno, R.~A. Newcombe, H.~Strasdat, P.~H. Kelly, and A.~J.
  Davison, ``{SLAM++}: Simultaneous localisation and mapping at the level of
  objects,'' {\em IEEE Int. Conf. on Computer Vision and Pattern Recognition
  (CVPR)}, 2013.

\bibitem{hermans14icra}
A.~Hermans, G.~Floros, and B.~Leibe, ``Dense {3D} semantic mapping of indoor
  scenes from rgb-d images,'' in {\em IEEE Int. Conf. on Robotics and
  Automation (ICRA)}, pp.~2631--2638, 2014.

\bibitem{stueckler15_semslam}
J.~St\"uckler, B.~Waldvogel, H.~Schulz, and S.~Behnke, ``Dense real-time
  mapping of object-class semantics from {RGB-D} video,'' {\em J. of Real-Time
  Image Processing}, 2015.

\bibitem{armeni16cvpr}
I.~Armeni, O.~Sener, A.~R. Zamir, H.~Jiang, I.~Brilakis, M.~Fischer, and
  S.~Savarese, ``{3D} semantic parsing of large-scale indoor spaces,'' in {\em
  IEEE Computer Vision and Pattern Recognition (CVPR)}, 2016.

\bibitem{riegler2016_octnet}
G.~Riegler, A.~O. Ulusoy, and A.~Geiger, ``{OctNet}: Learning deep 3d
  representations at high resolutions,'' {\em CoRR}, vol.~abs/1611.05009, 2016.

\bibitem{mccormac2016_semanticfusion}
J.~McCormac, A.~Handa, A.~J. Davison, and S.~Leutenegger, ``{SemanticFusion}:
  Dense {3D} semantic mapping with convolutional neural networks,'' {\em CoRR},
  vol.~abs/1609.05130, 2016.

\bibitem{whelan16_elasticfusion}
T.~Whelan, R.~F. Salas-Moreno, B.~Glocker, A.~J. Davison, and S.~Leutenegger,
  ``{ElasticFusion}: Real-time dense {SLAM} and light source estimation,'' {\em
  Intl. J. of Robotics Research (IJRR)}, 2016.

\bibitem{he2017_std2p}
Y.~He, W.~Chiu, M.~Keuper, and M.~Fritz, ``{STD2P}: Rgbd semantic segmentation
  using spatio-temporal data driven pooling,'' in {\em IEEE Int. Conf. on
  Computer Vision and Pattern Recognition (CVPR)}, 2017.

\bibitem{kundu2016_featurespaceoptim}
A.~Kundu, V.~Vineet, and V.~Koltun, ``Feature space optimization for semantic
  video segmentation,'' in {\em IEEE Int. Conf. on Computer Vision and Pattern
  Recognition (CVPR)}, 2016.

\bibitem{su2015_mvcnn}
H.~Su, S.~Maji, E.~Kalogerakis, and E.~G. Learned{-}Miller, ``Multi-view
  convolutional neural networks for 3d shape recognition,'' in {\em IEEE Int.
  Conf. on Computer Vision (ICCV)}, 2015.

\bibitem{lee15aistats}
C.~Lee, S.~Xie, P.~W. Gallagher, Z.~Zhang, and Z.~Tu, ``Deeply-supervised
  nets,'' in {\em Proc. of the 18th Int. Conf. on Artificial Intelligence and
  Statistics (AISTATS)}, 2015.

\bibitem{dosovitskiy15iccv}
A.~Dosovitskiy, P.~Fischer, E.~Ilg, P.~Hausser, C.~Hazirbas, V.~Golkov,
  P.~van~der Smagt, D.~Cremers, and T.~Brox, ``Flownet: Learning optical flow
  with convolutional networks,'' in {\em The IEEE Int. Conf. on Computer Vision
  (ICCV)}, December 2015.

\bibitem{zeiler2013_stochasticpooling}
M.~Zeiler and R.~Fergus, {\em Stochastic pooling for regularization of deep
  convolutional neural networks}.
\newblock 2013.

\bibitem{kerl13iros}
C.~Kerl, J.~Sturm, and D.~Cremers, ``Dense visual {SLAM} for {RGB-D} cameras,''
  in {\em IEEE/RSJ Intelligent Robots and Systems (IROS)}, 2013.

\bibitem{jaderberg15nips}
M.~Jaderberg, K.~Simonyan, A.~Zisserman, and k.~kavukcuoglu, ``Spatial
  transformer networks,'' in {\em Advances in Neural Information Processing
  Systems (NIPS)} (C.~Cortes, N.~D. Lawrence, D.~D. Lee, M.~Sugiyama, and
  R.~Garnett, eds.), 2015.

\bibitem{silberman12eccv}
P.~K. Nathan~Silberman, Derek~Hoiem and R.~Fergus, ``Indoor segmentation and
  support inference from {RGBD} images,'' in {\em Europ. Conf. on Computer
  Vision (ECCV)}, 2012.

\bibitem{gupta13cvpr}
S.~Gupta, P.~Arbelaez, and J.~Malik, ``Perceptual organization and recognition
  of indoor scenes from {RGB-D} images,'' in {\em IEEE Conf. on Computer Vision
  and Pattern Recognition (CVPR)}, 2013.

\bibitem{jia14caffe}
Y.~Jia, E.~Shelhamer, J.~Donahue, S.~Karayev, J.~Long, R.~Girshick,
  S.~Guadarrama, and T.~Darrell, ``Caffe: Convolutional architecture for fast
  feature embedding,'' {\em arXiv preprint arXiv:1408.5093}, 2014.

\bibitem{simonyan14vgg}
K.~Simonyan and A.~Zisserman, ``Very deep convolutional networks for
  large-scale image recognition,'' {\em CoRR}, vol.~abs/1409.1556, 2014.

\bibitem{he15iccv}
K.~He, X.~Zhang, S.~Ren, and J.~Sun, ``Delving deep into rectifiers: Surpassing
  human-level performance on imagenet classification,'' in {\em IEEE Int. Conf.
  on Computer Vision (ICCV)}, 2015.

\bibitem{bottou12sgd}
L.~Bottou, ``Stochastic gradient descent tricks,'' in {\em Neural Networks:
  Tricks of the Trade}, pp.~421--436, Springer, 2012.

\bibitem{ankur16cvpr}
A.~Handa, V.~Patraucean, V.~Badrinarayanan, S.~Stent, and R.~Cipolla,
  ``Scenenet: Understanding real world indoor scenes with synthetic data,'' in
  {\em IEEE Comp. Vision and Pattern Recognition (CVPR)}, 2016.

\bibitem{krahenb11nips}
P.~Kr\"{a}henb\"{u}hl and V.~Koltun, ``Efficient inference in fully connected
  {CRFs} with {G}aussian edge potentials,'' in {\em Advances in Neural
  Information Processing Systems (NIPS)}, 2011.

\end{thebibliography}

\end{document}